\documentclass{svproc}
\usepackage{url}
\usepackage{hyperref}
\usepackage{graphicx}
\usepackage{amssymb}
\usepackage{amsmath}
\usepackage{hyphenat}
\usepackage{listings}
\usepackage{placeins}
\usepackage{tabularx}
\newcolumntype{Y}{>{\centering\arraybackslash}X}

\lstdefinestyle{wraptt}{
  basicstyle=\ttfamily\small,
  breaklines=true,
  breakatwhitespace=false,
  columns=fullflexible,
  keepspaces=true,
  upquote=true
}

\begin{document}
\mainmatter              % start of a contribution
\title{VCBench: Benchmarking LLMs in Venture Capital}
%
  % abbreviated title (for running head)

\author{Rick Chen\inst{1} \and Joseph Ternasky\inst{2} \and
Afriyie Samuel Kwesi\inst{2} \and Ben Griffin\inst{1} \and Aaron Ontoyin Yin\inst{2} \and Zakari Salifu\inst{2} \and Kelvin Amoaba\inst{2} \and Xianling Mu\inst{1} \and Fuat Alican\inst{2} \and Yigit Ihlamur\inst{2}}
\authorrunning{Chen et al.} % abbreviated author list (for running head)
\tocauthor{First Author, Second Author, Third Author and Fourth Author}
\institute{University of Oxford\\
\email{rick.chen@seh.ox.ac.uk}
\and
Vela Research\\
\email{yigit@vela.partners}
}

\maketitle              % typeset the title of the contribution

\begin{abstract}
Benchmarks such as SWE-bench and ARC-AGI demonstrate how shared datasets accelerate progress toward artificial general intelligence (AGI). We introduce \textbf{VCBench}, the first benchmark for predicting founder success in venture capital (VC), a domain where signals are sparse, outcomes are uncertain, and even top investors perform modestly. At inception, the market index achieves a precision of 1.9\%. Y Combinator outperforms the index by a factor of 1.7×, while tier-1 firms are 2.9× better. \textbf{VCBench} provides 9,000 anonymized founder profiles, standardized to preserve predictive features while resisting identity leakage, with adversarial tests showing more than 90\% reduction in re-identification risk. We evaluate nine state-of-the-art large language models (LLMs). DeepSeek-V3 delivers over 6× the baseline precision, GPT-4o achieves the highest $F_{0.5}$, and most models surpass human benchmarks. Designed as a public and evolving resource available at \href{https://vcbench.com}{vcbench.com}, \textbf{VCBench} establishes a community-driven standard for reproducible and privacy-preserving evaluation of AGI in early-stage venture forecasting.
% We would like to encourage you to list your keywords within
% the abstract section using the \keywords{...} command.
\end{abstract}
\section{Introduction}
Benchmark datasets have played a defining role in the progress of machine learning (ML). By turning open-ended challenges into standardized and measurable tasks, they have enabled reproducible comparisons and driven entire fields forward. As models advance, there is growing demand for benchmarks that not only test raw accuracy but also allow systematic comparisons between machine and human performance. 

Venture capital (VC) is a compelling testbed for evaluating expert forecasting, offering a real-world setting to measure whether models can match or surpass human reasoning under uncertainty. Decisions rely on sparse and uncertain signals from founder backgrounds and early company data, while the financial stakes are high. At the inception and seed stages, even leading investors perform modestly in identifying the most successful startups -- those that were acquired or had an initial public offering (IPO) with a valuation exceeding \$500M, or raised more than \$500M in funding. The market index achieves 1.9\% precision at inception, while Y Combinator reaches 3.2\% ($1.7\times$ the index) and tier-1 VC firms are at 5.6\% ($2.9\times$). Recent models \cite{mu25,griffin2025} show that founder profiles alone can yield strong predictive signals, but the field lacks a standardized benchmark. These limits highlight both the practical importance and the scientific value of VC. It provides a domain where forecasting skill can be measured against expert performance, and exceeding that benchmark would represent a step toward human-level decision-making and AGI.

Building a benchmark for VC faces unique challenges. Founder data is abundant but noisy. LinkedIn provides wide coverage, yet its entries are inconsistent and unregulated. Crunchbase offers structured business metrics, but with limited scope. Together, these sources create four key issues: (i) \textit{format irregularity}, where the same degree or role appears in many variants; (ii) \textit{entry irregularity}, such as inclusion of internships or short courses; (iii) \textit{coverage imbalance}, with many missing or incomplete fields; and (iv) \textit{data contamination}, where large language models (LLMs) can re-identify founders from profile text and bypass the intended prediction task. These challenges differ from those in previous benchmarks and require a different approach.

In this paper we introduce \textbf{VCBench}, a standardized and anonymized benchmark for founder-success prediction. The dataset contains 9,000 founder profiles, with 810 labeled as successful. A key challenge is ensuring that models cannot bypass the prediction task by re-identifying founders. To address this, we propose and conduct adversarial re-identification experiments, which reduce identifiable founders by 92\% while preserving predictive features. Building on this foundation, we design a pipeline of standardization, filtering, enrichment, and anonymization. We then evaluate the performance of nine state-of-the-art LLMs on VCBench. To encourage further study, we release a public leaderboard for continuous evaluation of both LLMs and purpose-built VC models.

\subsubsection{Our contributions.}
\begin{enumerate}
    \item We introduce \textbf{VCBench}, the first anonymized benchmark for founder-success prediction in VC.  
    \item We develop a generalizable pipeline for data cleaning and anonymization, and validate it with adversarial re-identification tests. 
    \item We benchmark nine state-of-the-art LLMs, several surpassing human baselines, and release a public leaderboard at \href{https://vcbench.com}{vcbench.com}.  
\end{enumerate}

VCBench's novelty lies not only in its domain but also in the methodological design of its data standardization and anonymization pipelines, which address the distinctive challenges in the VC context. Together, our contributions represent an experimental step toward AGI benchmarking in identification-sensitive and history-dependent domains.

\subsubsection{Organization of this paper.} We first review the related literature in Section~2. Then Section~3 introduces the dataset and defines the success criterion. Section~4 presents our novel methodology for constructing the benchmark, followed by Section~5, which explains the adversarial anonymization tests in detail. Section~6 presents the VCBench leaderboard and LLM evaluation results. Finally, discussions and concluding remarks are included in Sections~7 and 8.

\section{Related Work}

\subsubsection{Benchmarks in machine learning.}
Benchmark datasets have long been central to machine learning progress. In vision and language, VQA \cite{VQA} and ImageNet \cite{ImageNet} defined early standards, while in reasoning and problem solving, ARC-AGI \cite{ARC-AGI,ARC-AGI2} has become a touchstone. Applied domains have also introduced specialized benchmarks, such as SDBench for medical diagnosis \cite{SDBench} and SWE-bench for software engineering \cite{SWE}. These benchmarks share the common goal of translating an open-ended task into a reproducible and measurable challenge that accelerates research. However, constructing such benchmarks in the VC domain is considerably more challenging, as outlined in the introduction and presented in Table~\ref{table:comparisons_benchmarks}. In particular, unlike medical data \cite{SDBench}, founder information is by default published without anonymization; while anonymization mitigates re-identification risks, it usually destroys valuable predictive features simultaneously. To our knowledge, no existing benchmark evaluates decision-making under uncertainty in socioeconomic contexts such as VC. \textbf{VCBench} fills this gap by establishing a standardized and anonymized testbed for early-stage startup success prediction.

\begin{table*}[h]
\centering
\caption{Compared with existing LLM benchmarks, the construction of VCBench faces a unique combination of challenges}
\begin{tabularx}{\textwidth}{lYYYY}
\hline
\textbf{Benchmark}            & \textbf{ARC-AGI-2} & \textbf{SWE-Bench} & \textbf{Microsoft SDBench} & \textbf{VCBench} \\ \hline
\textbf{Data Filtering}       & --                 & \checkmark         & --                         & \checkmark       \\ \hline
\textbf{Data Standardization} & --                 & --                 & --                         & \checkmark       \\ \hline
\textbf{Coverage Improvement} & --                 & --                 & \checkmark                 & \checkmark       \\ \hline
\textbf{Data Anonymization}   & --                 & --                 & --                         & \checkmark       \\ \hline
\end{tabularx}

\label{table:comparisons_benchmarks}
\end{table*}

\subsubsection{Venture capital prediction.}
Many studies have explored predictive models for startup success, utilizing a variety of ML techniques and data formats \cite{dellermann2021findingunicornpredictingearly,RazaghzadehBidgoli2024,potanin2023startupsuccesspredictionvc,Maarouf_2025,Javier2019}. Closely related works, Mu et al. \cite{mu25} and Griffin et al. \cite{griffin2025}, show that founder profiles alone can yield strong predictive signals. Methodologies in the domain vary widely, ranging from traditional statistical techniques such as Random Forests and clustering algorithms to modern LLM-powered frameworks. However, these studies rely on proprietary or nonstandard datasets, limiting reproducibility and comparability. Addressing major challenges such as data irregularity and data contamination is essential for constructing any public benchmark. VCBench provides the first standardized and anonymized dataset in VC that enables fair comparison across models and establishes human-level baselines for evaluation.

\subsubsection{Data contamination.}
A growing body of research highlights risks of data contamination/leakage in LLM evaluation \cite{BDCSurvey2024,AntiLeakBench,RethinkContamination}. Models may recognize entities from their pre-training corpora and bypass intended reasoning tasks by re-identification. In the VC setting, this problem is particularly acute, as founder profiles are often public. However, most prior studies address contamination in question-answering benchmarks, where entire question–answer pairs may appear in training data. There, semantic detection and data-refreshing strategies are effective. In contrast, re-identifying founders requires cross-referencing multiple noisy sources rather than simple memorization, making such methods less reliable. Moreover, founder-success prediction relies on long historical horizons (8–12 years), rendering data-refresh approaches impractical without synthetic augmentation.

To our knowledge, no existing analysis is based on a real-world decision-making setting like VC. To address this challenge, we propose a multistage anonymization pipeline validated by adversarial re-identification experiments, achieving a 92\% reduction in identifiable founders in VCBench while preserving predictive features. A key advantage of our pipeline is that it ensures faithful anonymization by not including data synthesization or noise injection. In addition, we follow the modern benchmarking practice \cite{ARC-AGI,BDCSurvey2024} of keeping the evaluation set private to mitigate future leakage risks.

\subsubsection{Summary.}
In contrast to previous benchmarks, predictive models and anonymization studies, VCBench integrates these threads into a single resource. It offers the first standardized benchmark for founder-success prediction, designed to evaluate models fairly against human expertise while preventing identity leakage.

\section{Dataset}

\subsubsection{Scale and success definition.}
VCBench contains 9,000 anonymized founder profiles with a 9\% baseline success rate (810 founders). Each founder is paired with their most recently founded company, which determines the success label. A founder is considered successful if the company was acquired or had an initial public offering (IPO) with a valuation exceeding \$500M, or raised more than \$500M in funding. Such cases are typically known as outliers. Companies that raised \$100K--\$4M at inception but did not achieve an exit, IPO, or substantial follow-on funding within eight years of founding are classified as unsuccessful. The eight-year horizon reflects expert practice: it typically takes 8--12 years for startups to reach an IPO, a large-scale funding, or an exit. Those that fail to do so are not considered outlier successes in VC.

\subsubsection{Coverage.}
Most companies were founded in the U.S. between 2010 and 2018. In this period, roughly 5,000 U.S. startups raised more than \$100K annually, suggesting a base population of 45,000. \textbf{VCBench} therefore covers about a fifth of this population, making it statistically representative of the early-stage startup landscape in the U.S. Distributions of industries and founding years are presented in Fig.~\ref{fig:industry} and Fig.~\ref{fig:startyear} in Appendix \ref{app:distributions}. The U.S. is chosen as the focus geography because it represents the largest entrepreneurship ecosystem globally.

\subsubsection{Fields.}
Each founder record contains structured fields capturing founder attributes and company outcomes:
\begin{itemize}
    \item \texttt{success}: Binary label indicating founder success as defined above.  
    \item \texttt{industry}: Sector of the startup.  
    \item \texttt{ipos} and \texttt{acquisitions}: Details of the founder's prior experience in IPOs and exits as a co-founder.  
    \item \texttt{educations\_json}: List of education records with \texttt{degree}, \texttt{field}, and \texttt{QS ranking}.  
    \item \texttt{jobs\_json}: List of job records with \texttt{role}, \texttt{company size}, \texttt{industry}, and \texttt{duration}.  
\end{itemize}

\subsubsection{Sources and filtering.}
Founder and company data was collected from licensed and publicly available data sources. LinkedIn provides comprehensive education and job histories. Crunchbase helps with enrichment and contributes IPO, acquisition, and funding records. For each founder, information is restricted to data available prior to the founding of the company, simulating real-world early-stage prediction.

\subsubsection{Formats.}
To accommodate both LLMs and ML models, VCBench is distributed in two complementary formats. The first is \texttt{anonymized\_prose}, a natural-language summary of each founder constructed from structured fields, designed for direct LLM input. The second format exposes structured JSON fields, enabling feature-level access for custom ML models. 

\section{Methodology}
\label{sec:methodology}

\begin{figure*}[t]
    \centering
    \includegraphics[width=0.95\linewidth]{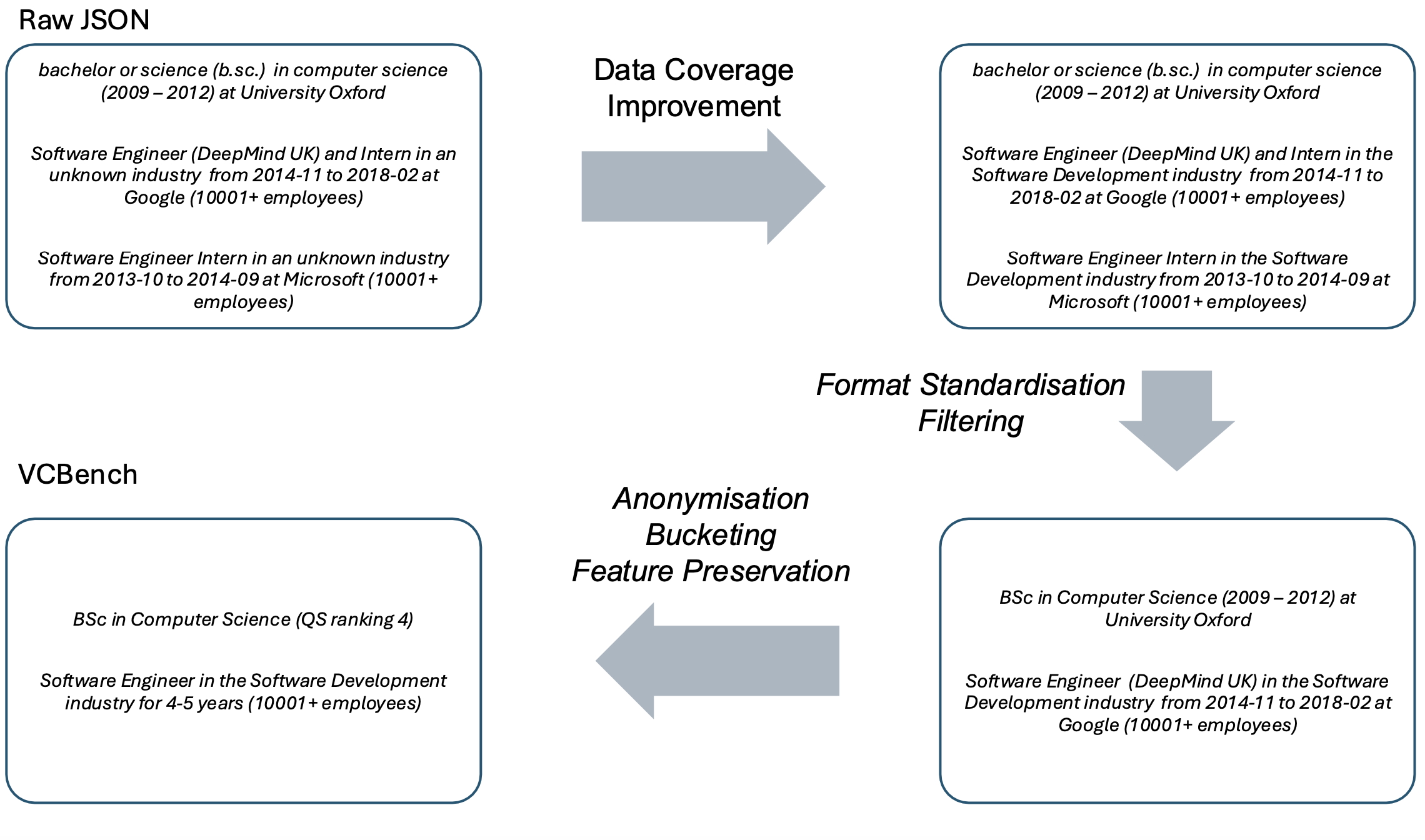}
    \caption{Data Cleaning Pipeline}
    \label{fig:data_cleaning_pipeline}
\end{figure*}

We construct VCBench in four stages: (i) data coverage improvement, (ii) format standardization and filtering, (iii) anonymization at both the entry and dataset levels, and (iv) iterative field selection guided by adversarial testing. Fig.~\ref{fig:data_cleaning_pipeline} visualizes the full pipeline on an example profile, while Table~\ref{tab:standardisation_and_filtering} summarizes vocabulary compression after standardization and filtering.

\subsection{Data Coverage Improvement}
LinkedIn offers wide coverage but irregular entries. Crunchbase is structured yet incomplete. Empty fields can obscure the signal. For example, an empty acquisition field may indicate the absence of exits or missing information. To mitigate this without speculative inference, we adopt two steps:
\begin{enumerate}
    \item \textbf{Data source cross-checking.} When a field is empty in one source, we retrieve the corresponding value from the other source when a direct match is apparent.
    \item \textbf{Cross-record consistency.} We enforce consistent values for the same entity across profiles (e.g., the industry label attached to the same organization) to reduce label drift.
\end{enumerate}

These steps reduce irregularities and missing values while maintaining a conservative stance toward uncertain or conflicting entries.

\subsection{Format Standardization and Data Filtering}

LinkedIn entries contain spelling variants, inconsistent punctuation, and heterogeneous representations of degrees and roles (e.g., ``PhD'', ``p.h.d.'', ``Doctor of Philosophy''). We adopt a two-step process:
\begin{enumerate}
    \item \textbf{Deterministic canonicalization:} Trim whitespace, normalize conjunctions (``and'', ``\&'', ``/''), punctuation, and common aliases for degrees and roles.
    \item \textbf{LLM-assisted flagging:} LLM identifies strings requiring reformatting and tags nonformal entries using explicit exclusion categories (e.g., ``Intern'', ``Course'', ``Visiting'').
\end{enumerate}

Through this process, non-formal education and job entries are flagged and removed to avoid systematic bias. After filtering, founders with no remaining job record are excluded from the dataset to ensure well-formed professional histories. Overall, this stage reduces label fragmentation and consolidates noisy vocabularies while preserving predictive structure (Table~\ref{tab:standardisation_and_filtering}).

\begin{table*}[t]
\centering
\caption{An entry-level summary of the format standardization and data filtering process}

\begin{tabularx}{\textwidth}{lYYYY}
\hline
Record Type   & Original No.\ unique entries & Final No.\ unique entries & Percentage Reduction \\
\hline
industry & 314 & 61 &  80.6\% \\
education degree & 2155 & 404 & 81.3\% \\
education field of study & 6360 & 3969 & 37.6\% \\
job role & 21259 & 16374 & 23.0\% \\
education record & 20573 & 15620 & 24.1\% \\
job record & 45975 & 41183 & 10.4\% \\
\hline
\end{tabularx}
\label{tab:standardisation_and_filtering}
\end{table*}

\subsection{Anonymization}
We consider three adversary classes by increasing identification capability: general-purpose LLMs (e.g., GPT-4o), reasoning models (e.g., o3), and tool-assisted models with web search. Our approach combines entry-level and dataset-level anonymization.

\subsubsection{Entry-level anonymization.} We remove founder names, company names, locations, and dates from all fields, including nested text in \texttt{educations\_json} and \texttt{jobs\_json}. This prevents direct string lookup (e.g., job titles such as ``Software Engineer (Microsoft India)'').

\subsubsection{Dataset-level anonymization.} To reduce linkage via rare attribute combinations \cite{kAnonymity}, we group simple numeric fields (\texttt{ipos} and \texttt{acquisitions}). For \texttt{industry}, we follow the following procedure:
\begin{enumerate}
    \item Embed industry labels using an OpenAI embedding model.
    \item Apply agglomerative hierarchical clustering with cosine similarity.
    \item Refine clusters using o3 reasoning.
    \item Perform a final human pass.
\end{enumerate}
This yields 61 clusters, each containing at least 10 founders.

Education prestige is preserved using QS university rankings, with aliases handled by an LLM to unify naming conventions (e.g., ``California Institute of Technology'' vs.\ ``Caltech''). Job start and end dates are converted into duration buckets, expressed in years, which preserve career trajectory information while concealing exact timelines.

\subsection{Iterative Anonymization and Feature Selection Process}

We refine the dataset with \emph{anonymization unit tests} that explicitly target founder identification rather than success prediction. Initial experiments (Appendix \ref{app:initial_experiments}) using the o3 model due to its powerful reasoning power (with and without web search) identified leakage pathways such as specific institution names, company names, funding/IPO/acquisition values, and exact dates. We therefore removed or bucketed these fields and re-tested as described in the prior sections.

At each iteration, we adopted the following loop: propose a change, run identification tests, and retain the change only if it reduces re-identification while preserving predictive structure. Founders who were identified on at least two occasions are removed to reduce the risk of data contamination. 

\section{Experiments and Ablation Analysis}

The purpose of the experiments is to validate whether VCBench achieves strong anonymization while preserving predictive signal. We design \emph{anonymization unit tests} in which models are explicitly instructed to re-identify founders rather than predict success. This setting provides a direct measure of the residual leakage risk.

\subsubsection{Experimental design.} We sample $300$ successful founders, as these represent the most high-profile cases and thus the hardest anonymization challenge. For each profile, we test whether an adversary could correctly infer the founder's identity. We employ two adversaries representing distinct threat models: DeepSeek-R1, a reasoning model without web access (offline), and Gemini-2.5-Pro with grounding (web-search, online). The evaluation metric is the proportion of correctly identified founders.

\subsubsection{Ablation formats.} We compare five input formats, each progressively modifying the representation:
\begin{itemize}
    \item \textbf{JSON:} structured education and job records before anonymization but excluding IPOs and acquisitions. 
    \item \textbf{Base (anonymized JSON):} The anonymized format with the founder names and institution names removed, adding IPOs and acquisitions.
    \item \textbf{Base + QS rankings (bucketed):} The anonymized format enriched with QS ranks grouped into intervals (e.g. 1-20, 20-100).
    \item \textbf{Base + QS rankings (unbucketed):} The anonymized format enriched with exact QS ranks. 
    \item \textbf{Final Format (VCBench):} The final anonymization format with unbucketed QS ranks and bucketed job durations.
\end{itemize}

\subsubsection{Results.} Identification rates are reported in Table~\ref{table:anonymisation_unit_testing}. Moving from the pre-anonymization JSON to the final format reduced online identification by approximately $80\%$ and offline identification by $92\%$. An unexpected but beneficial finding was that explicit QS rankings reduced identification rates further. Reasoning logs showed that models often misused current QS lists, which in turn blurred re-identification. This allowed us to preserve educational prestige while strengthening anonymity.

\begin{table}[h] 
\centering
\caption{Anonymization unit testing results: average identification rates for different input formats. Bold formats were tested three times for reliability}
\begin{tabular}{l c c c}
\hline
Input & Online \% & Offline \% \\
\hline
\textbf{Processed JSON} &  \textbf{77.0} & \textbf{17.2} \\
Base &  18.3 & 1.2 \\
QS (bucketed) & 15.4 & 2.3 \\
QS (unbucketed) & 12.7 & -- \\
\textbf{Final Format (VCBench)} & \textbf{15.1} & \textbf{1.3} \\
\hline
\end{tabular}

\label{table:anonymisation_unit_testing}
\end{table}

\subsubsection{Representativeness.} Our sample of 300 covers more than a third of the successful founder cohort, enhancing the generalizability of our results to the broader successful population. Moreover, small-scale experiments that we ran suggest that the unsuccessful founders are about three times less likely to be identified, supporting our decision to monitor and control the level of anonymization of successful founders.

\subsubsection{Interpretation.} By removing identifiers, bucketing rare values, and clustering industries, multistage anonymization reduces leakage even under strong adversaries. The final VCBench format balances data contamination with feature diversity, supporting robust benchmarking and fair comparison.

\section{The VCBench Leaderboard}

After constructing VCBench, we evaluated the predictive performance of nine state-of-the-art LLMs. The dataset of 9{,}000 founders was split into six folds of 1{,}500 each, preserving the 9\% success rate. Performance was measured using the $F_{0.5}$ score, which weights precision twice as heavily as recall:

\begin{equation}
    F_{0.5} = (1 + 0.5^2) \cdot \frac{\text{Precision} \cdot \text{Recall}}{(0.5^2 \cdot \text{Precision}) + \text{Recall}}.
\end{equation}

This emphasis reflects the VC setting, where false positives are more costly than false negatives. The average results are reported in Table~\ref{tab:leaderboard_performance}, with fold-specific outcomes in Appendix \ref{app:fold_specific}.

\subsubsection{Results.} GPT-4o achieved the highest $F_{0.5}$ (25.1\%). DeepSeek-V3 recorded the highest precision (59.1\%) but with a low recall, while Gemini-2.5-Flash obtained the highest recall (69.1\%) but modest precision.  
\begin{table}[t]
\centering
\caption{Predictive performance of nine LLMs on VCBench, averaged across six folds and ordered by average $F_{0.5}$. Results should be interpreted cautiously when extrapolating from the 9\% dataset success rate to the real-world 1.9\% success rate }
\setlength{\tabcolsep}{4pt}
\begin{tabular}{l c c c}
\hline
Model & Precision (\%)& Recall (\%)& $F_{0.5}$ (\%)\\
\hline
GPT-4o              & 29.1 & 16.2 & \textbf{25.1} \\
DeepSeek-R1         & 37.6 &  8.4 & 22.1 \\
GPT-4o-mini         & 29.5 & 10.1 & 21.2 \\
o3                  & 42.4 &  7.0 & 20.9 \\
Gemini-2.5-Pro      & 17.2 & 59.0 & 20.1 \\
Claude-3.5-Haiku    & 16.9 & 48.6 & 19.4 \\
GPT-5               & 53.7 &  4.3 & 16.2 \\
Gemini-2.5-Flash    & 12.6 & \textbf{69.1} & 15.1 \\
DeepSeek-V3         & \textbf{59.1} &  3.0 & 11.8 \\
\hline
\end{tabular}

\label{tab:leaderboard_performance}
\end{table}

\subsubsection{Cost and efficiency.} Inference cost and latency are summarized in Table~\ref{tab:leaderboard_efficiency}. GPT-4o-mini offers strong performance at low cost, while DeepSeek-V3 offers cost-effective reasoning with better performance over other reasoning models such as Gemini-2.5-Pro and GPT-5. 

\subsubsection{Leakage check.} The comparatively high precision obtained on fold~4 documented in Appendix \ref{app:fold_specific} (Table \ref{tab:per_fold_results}) motivated closer inspection. We reviewed the prediction reasoning logs and ran additional anonymization unit tests on a sample of 300 founders in that fold. Identification rates remained low (4.3\% for o3; 7.3\% for Gemini-2.5-Pro with web search), and no explicit identifications appeared in model outputs. The elevated precision was instead due to a concentration of founders with extreme outcomes (e.g., prior IPO/acquisition events with valuations exceeding \$500M), confirming that improvements reflected predictive signal rather than identity leakage.

\subsubsection{Interpretation.} 
\begin{figure}[h]
    \centering
    \includegraphics[width=0.7\textwidth]{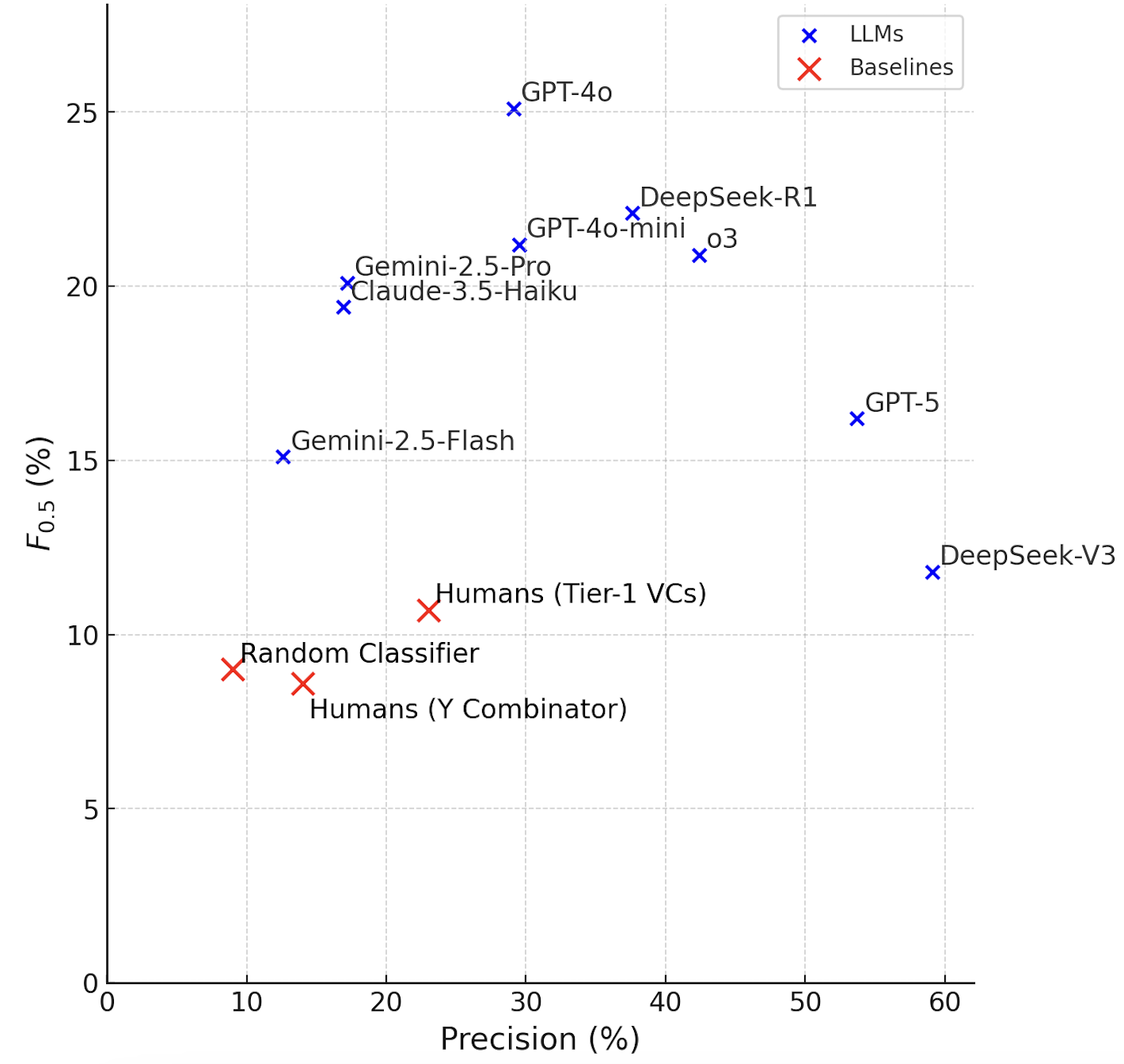}
    \caption{$F_{0.5}$-score versus precision for nine vanilla LLMs on VCBench. Human-baseline results are scaled linearly to  reflect the inflation of success rate from the real-world (1.9\%) to VCBench (9\%) }
    \label{fig:leaderboard}
\end{figure}

These results demonstrate that anonymized founder profiles preserve enough predictive signal for LLMs to outperform human experts in startup investing. GPT-4o achieves the best $F_{0.5}$ score (25.1\%) with a precision of 29\%, representing a 3.2$\times$ improvement over the precision baseline and exceeding the 2.9$\times$ performance of tier-1 VC firms (Fig. \ref{fig:leaderboard}). DeepSeek and Gemini models highlight different points on the precision--recall frontier: DeepSeek-V3 produces highly precise predictions with low coverage, whereas Gemini-2.5-Flash recalls nearly 70\% of successful founders at the expense of precision. Crucially, leakage tests confirm that these gains are not explained by identity re-identification. We also report ablation results in Appendix~\ref{app:pre-anonymization_rates}, where pre-anonymization inputs are used for comparison.

\subsubsection{Mitigation of Future Leakage.}
An essential consideration for modern benchmarks is mitigating the risk of leakage into the pre-training corpus of future LLMs, which would introduce inherent biases into model comparisons. We address this by only releasing half of the dataset (4{,}500 founders) publicly. The remaining half is reserved for private evaluation. Leaderboard scores are computed on the private fold, preserving the integrity of model assessments.

\begin{table}[t]
\centering
\caption{Cost per $1$M tokens and latency for models evaluated on VCBench, ordered by average $F_{0.5}$}
\begin{tabular}{l c c}
\hline
Model & Cost In/Out & Latency \\
\hline
GPT-4o              & \$2.50 / \$10.00 & 3.59 s \\
DeepSeek-R1         & \$0.55 / \$2.19  & 37.83 s \\
GPT-4o-mini         & \$0.15 / \$0.60  & 3.04 s \\
o3                  & \$2.00 / \$8.00  & 6.90 s \\
Gemini-2.5-Pro      & \$1.25 / \$10.00 & 10.73 s \\
Claude-3.5-Haiku    & \$0.80 / \$4.00  & 3.36 s \\
GPT-5               & \$1.25 / \$10.00 & 1.54 s \\
Gemini-2.5-Flash    & \$0.30 / \$2.50  & 8.32 s \\
DeepSeek-V3         & \$0.27 / \$1.10  & 10.07 s \\
\hline
\end{tabular}

\label{tab:leaderboard_efficiency}
\end{table}

\section{Discussion}
We assess \textbf{VCBench} in terms of benchmark validity and design trade-offs. Its interpretation is bounded by several factors: the inflated success rate (9\% vs.\ 1.9\% in the real-world), potential sampling bias from LinkedIn/Crunchbase, right-censoring from the eight-year horizon, and residual noise after standardization. To extend utility without weakening predictive value, we suggest anonymized proxies for company prestige, scalable clustering for high-cardinality fields, trajectory-level features that balance privacy and signal, broader model coverage, and a sequential simulation mode for decision-making under resource constraints. Together, these directions clarify how current scores should be read and outline a path toward a more realistic and robust benchmark.

\subsection{Limitations}
VCBench carries several known caveats:
\begin{itemize}
    \item \textbf{Prevalence shift.} The benchmark sets the success rate at 9\% rather than the real-world 1.9\%. This inflated prevalence stabilizes testing and improves statistical significance, but also introduces caveats. Precision does not scale linearly with prevalence, so relative multipliers (e.g., $3.2\times$ for GPT-4o versus $2.9\times$ for tier-1 VCs) may not hold under the true 1.9\% distribution. Reported gains should therefore be interpreted cautiously when extrapolated back to deployment settings.
    \item \textbf{Comparability of human baselines.} The normalization procedure assumes that human investors and LLMs face the same distribution of opportunities. In reality, VCs self-select their deal flow, and access is constrained by competition, reputation, and human bandwidth. This structural difference means that even after normalization, comparisons between model and human performance may misrepresent the actual gap in practice.
    \item \textbf{Bias in collected data.} Although VCBench covers a statistically significant share of U.S. founders from 2010 to 2018, it still inherits biases from LinkedIn and Crunchbase. Coverage is stronger for technology startups and founders with public visibility, while non-tech sectors, self-funded companies, and underrepresented demographics are more likely to be absent or inconsistently represented. These biases may skew the benchmark toward patterns that generalize poorly beyond the VC mainstream.
    \item \textbf{Temporal bias.} The eight-year horizon used to define success introduces a right-censoring effect. Companies founded later in the window are more likely to be labeled unsuccessful simply because their outcomes have not yet had time to materialize. For example, a startup founded in 2017 is penalized relative to one founded in 2012, which has had additional years to mature and demonstrate success. While the eight-year cutoff reflects the median time to major outcomes such as IPOs or acquisitions, it inevitably places more recent cohorts at a disadvantage.
    \item \textbf{Limited target population.} We focused on U.S.-based founders to reduce geographical bias, as data availability may vary significantly across regions. Extending the dataset to include other geographies like Europe would enhance the generalizability of the results and potentially reveal geographical differences.
    \item \textbf{Residual irregularities.} Despite multistage cleaning, large-scale founder data remains noisy. Stochastic LLM standardization and irregular entries make it difficult to guarantee full consistency.  
\end{itemize}

\subsection{Future Work}
Several directions can extend VCBench and address the limitations mentioned above. First, continued community-driven refinement will be important. Releasing updates as feedback arrives will improve data coverage and reduce residual irregularities, making VCBench a ``living benchmark'' rather than a static artifact. Second, anonymization can be enhanced by incorporating company prestige signals (e.g., Fortune-500 or NASDAQ-100) that preserve organizational context without revealing identities. Third, the current clustering approach works well for industries but not for high-cardinality fields like job roles or education. Developing scalable clustering methods would enable more complete dataset-level anonymization. 

On the feature side, more advanced feature engineering could transform structural and temporal patterns (such as job sequences, career trajectories, or prestige proxies) into predictive variables that reduce re-identification risk while strengthening model accuracy. With respect to evaluation, the leaderboard should expand to include both additional LLMs and domain-specific VC models, enabling richer comparisons across approaches. The integration of multimodal data, such as speech and video analysis, may also play a pivotal role in advancing toward end-to-end AGI in VC.

Finally, we envision a VC-simulation mode in a tournament-style format where startups appear sequentially and investment decisions must be made under budget constraints. This would capture the real dynamics of venture investing and test whether models can make consistent, resource-limited decisions. Looking further, we are exploring a \emph{gamified arena} where human investors, LLMs, and purpose-built VC models compete side by side, creating an open platform for benchmarking decision-making under uncertainty in real time.

\section{Conclusion}
We introduced \textbf{VCBench}, the first standardized and anonymized benchmark for founder-success prediction in venture capital. The dataset was constructed using a novel multistage anonymization pipeline, validated by adversarial tests, which reduced re-identification risk by over 90\% while preserving predictive signal. This addresses the unique challenges of building founder-based benchmarks, such as data irregularity and data contamination.

Using this benchmark, we evaluated nine state-of-the-art LLMs and found that several outperform not only the market index but also the leading VC firms, with GPT-4o achieving the highest $F_{0.5}$ score. These results show that anonymized founder profiles are sufficient to surpass human-expert baselines in early-stage venture forecasting.  

By releasing both the dataset and a public leaderboard, we provide a foundation for reproducible research in this high-stakes domain. In the future, we aim to address current limitations such as prevalence shift and dataset bias. \textbf{VCBench} is designed as a community-driven benchmark that will evolve with feedback, richer features, and new evaluation modes, including simulation and human--AI competitions, offering a path toward more realistic tests of decision-making under uncertainty.


\appendix
\section*{Appendix}
\section{Prompts and Input Formats}
\label{app:prompts}

\subsection{Example raw JSON founder profile}
\begin{lstlisting}[style=wraptt]
{
  "industry": "Research Services",
  "twitter_url": null,
  "jobs": [
    {
      "title": "Professor",
      "company": "Duke University",
      "company_industry": "Higher Education",
      "started_at": "2013-07-13",
      "ended_at": "still working"
    },
    {
      "title": "Founder, President, CEO",
      "company": "Applied Quantum Technologies",
      "company_industry": null,
      "started_at": "2006-08-13",
      "ended_at": "2020-02-13"
    },
  ],
  "educations": [
    {
      "university": "Young Dong High School",
      "degree": null,
      "fields": null,
      "started_on": null,
      "ended_on": null
    }
  ]
}
\end{lstlisting}

\subsection{Example anonymized prose format}
\begin{lstlisting}[style=wraptt]
"This founder leads a startup in the IT Services & Digital Solutions industry.
Education:
* BA in History (Institution QS rank 42)

Professional experience:
* Board Member for <2 years in the `Management, Strategy & Professional Services` industry (51-200 employees)
* Board Member for 2-3 years in the `Education & Training Services` industry (51-200 employees)
* Angel Investor, Advisor for 4-5 years
* Co-Founder, CEO for 6-9 years in the `Software Development` industry (51-200 employees)
* Co-Founder for <2 years

They have overseen one acquisition as a founder: one 15M - 50M in value."
\end{lstlisting}

\subsection{Anonymization unit testing prompt}
\begin{lstlisting}[style=wraptt]
You are given the profile of an entrepreneur who founded at least one startup after 2010. This includes information like the founder's education and job histories.

### Your Task
1. Examine the profile in JSON format below.
2. Deduce the most likely identity of the founder (using web search as an aid).
3. Respond in the exact format specified under "Required Output".

### OUTPUT FORMAT
Put the founder's full name right at the beginning. Avoid abbreviations or nicknames for first and last names. Do not include middle names. Do not include any other text before it.
Do not return the text "Founder's name". Always give a guess, even if you are not sure.
In addition, include the following information:
Reasoning: <one brief paragraph explaining how the data led you to this name>
Confidence: <Low, Medium, High>

### Inputs
{profile}
\end{lstlisting}

\section{VCBench Distributions}\label{app:distributions}

\begin{figure}[h]
    \centering
    \includegraphics[width=0.9\linewidth]{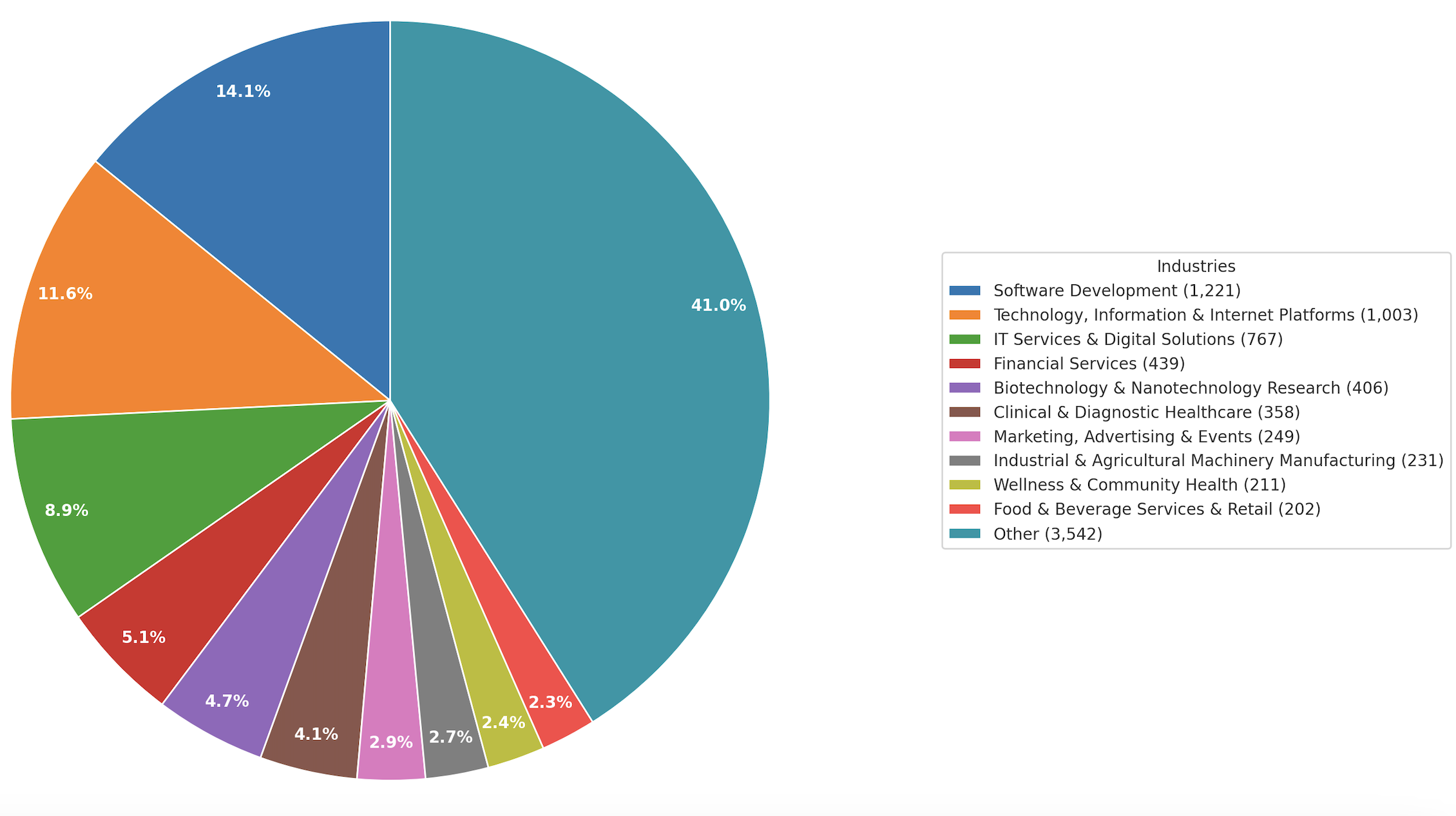}
    \caption{Distribution of industries in VCBench after bucketing}
    \label{fig:industry}
\end{figure}

\begin{figure}[h]
    \centering
    \includegraphics[width=0.9\linewidth]{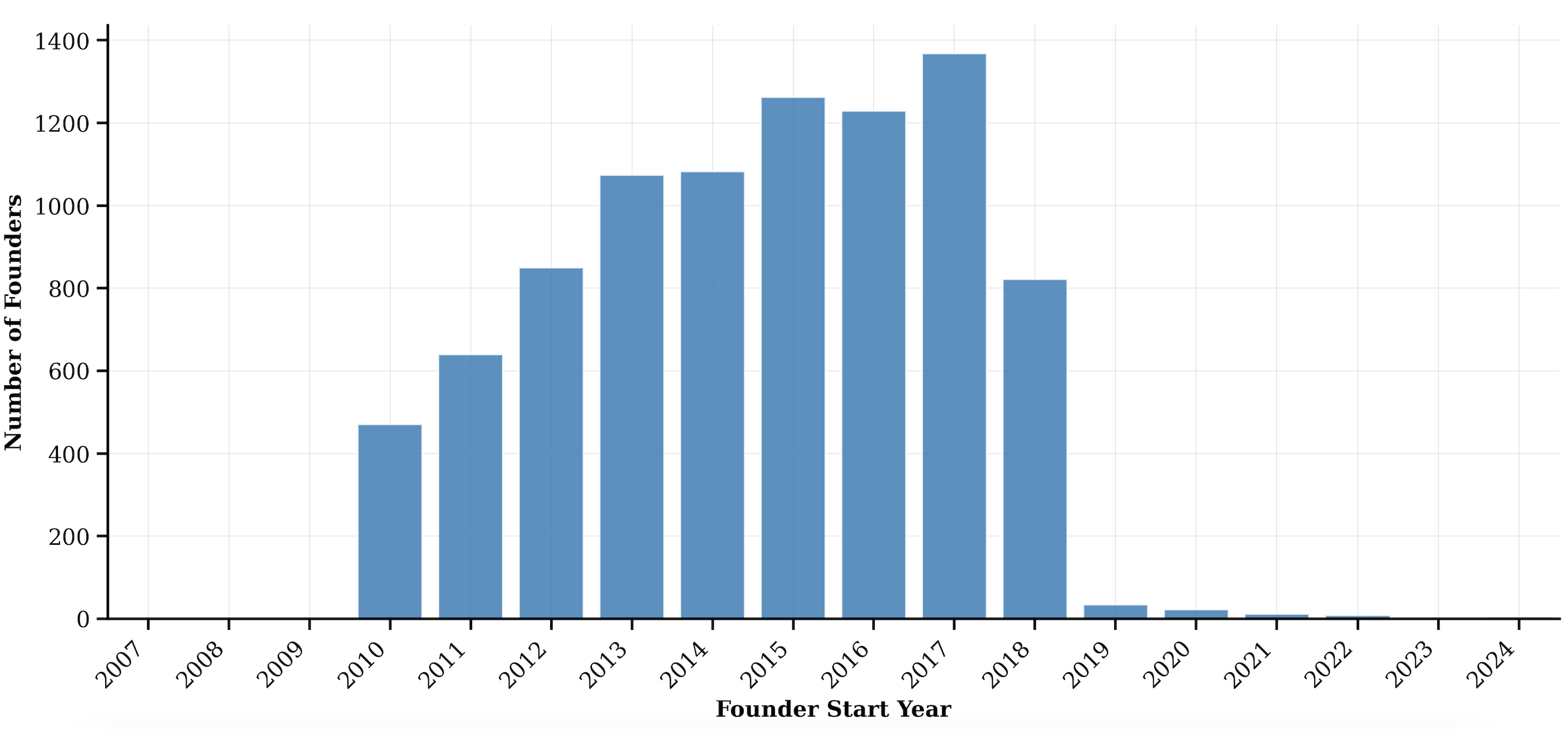}
    \caption{Distribution of startup founding years in VCBench}
    \label{fig:startyear}
\end{figure}

\FloatBarrier
\section{Example Initial Experiment with o3}\label{app:initial_experiments}
The following demonstrates one of the initial founder identification experiments. Using web searches, the OpenAI o3 model correctly identified the given founder in under two minutes, despite the absence of any explicit identifier. The exceptionally successful exit was the point of failure in this case.

\begin{lstlisting}[style=wraptt]

User:
You are an expert in venture capital tasked with identifying successful founders from their unsuccessful counterparts. All founders under consideration are sourced from LinkedIn profiles of companies that have raised between $100K and $4M in funding. A successful founder is defined as one whose company has achieved either an exit or IPO valued at over $500M. Given the following founder {'Industry': 'Biotechnology Research', 'Education': [], 'Work Experience': [{'title': 'VP Applications', 'company': 'consumer genomics unicorn (>$3 B exit)', 'company_industry': 'Biotechnology Research', 'started_at': '2010', 'ended_at': '2012'}, {'title': 'First Employee, Founding Architect, Director of R&D', 'company': 'digital PCR startup (>$150 M M&A exit)', 'company_industry': 'Biotechnology Research', 'started_at': '2006', 'ended_at': '2010'}]} 

Who is this founder?"


OpenAI o3:
I'm considering several possibilities for the "consumer genomics unicorn" with a >$3B exit. Ancestry.com was sold in 2012 for $1.6B, and later for $4.7B in 2020, which doesn't fit. The Grail spin-off from Illumina could work, but its exit didn't happen until 2020 with a $7B deal. Sequenom, a PCR startup, raised $150M for an exit around 2010 but was sold in 2017. Natera could also be a contender, though its valuation was smaller initially. Still, the Grail sale seems the closest.

(...)

The search results point to Serge Saxonov as matching both the QuantaLife and 23andMe roles, though the timeline is reversed in the data. This could be a misorder in the dataset, but the identity seems to be correct. I also considered Ben Hindson, though he didn't work at 23andMe. It looks like Serge Saxonov is the founder. Next step: let's confirm by checking for other people who may have had roles at both QuantaLife and 23andMe.
\end{lstlisting}

\section{Per-fold Results for Vanilla LLMs}\label{app:fold_specific}

\begin{table}[h]
\centering
\caption{Per-fold and average results for all models. All results are recorded as percentages. Note that DeepSeek-V3 encountered an undefined precision on Fold 3 (no positives predicted), so the average precision was calculated across the remaining 5 folds; average recall and $F_{0.5}$ were still calculated across all folds}
\begin{tabularx}{\textwidth}{lYYYYYYY}
\hline
Model & Fold 1 (\%)& Fold 2 (\%)& Fold 3 (\%)& Fold 4 (\%)& Fold 5 (\%)& Fold 6 (\%)& Avg (\%)\\
\hline
Claude-3.5-Haiku & & & & & & & \\
\quad Precision & 17.0 & 17.1 & 17.6 & 17.5 & 12.9 & \textbf{19.5} & 16.9 \\
\quad Recall    & 50.4 & 50.4 & 51.1 & 48.1 & 37.8 & \textbf{54.1} & 48.6 \\
\quad $F_{0.5}$ & 19.6 & 19.7 & 20.2 & 20.0 & 14.8 & \textbf{22.3} & 19.4 \\
\hline
DeepSeek-V3 & & & & & & & \\
\quad Precision & 77.8 & 50.0 & \textbf{--} & 42.9 & \textbf{83.3} & 41.7 & 59.1 \\
\quad Recall    & \textbf{5.2}  & 3.0  & \textbf{0.0} & 2.2  & 3.7  & 3.7  & 3.0 \\
\quad $F_{0.5}$ & \textbf{20.5} & 12.0 & \textbf{0.0} & 9.2  & 15.7 & 13.7 & 11.8 \\
\hline
DeepSeek-R1 & & & & & & & \\
\quad Precision & 27.0 & 30.0 & 22.2 & \textbf{50.0} & 46.2 & \textbf{50.0} & 37.6 \\
\quad Recall    & 7.4  & 6.7  & 4.4  & 11.1 & 8.9  & \textbf{11.9} & 8.4 \\
\quad $F_{0.5}$ & 17.7 & 17.6 & 12.4 & 29.4 & 25.1 & \textbf{30.4} & 22.1 \\
\hline
Gemini-2.5-Flash & & & & & & & \\
\quad Precision & \textbf{13.9} & 12.7 & 13.1 & 12.6 & 10.5 & 12.9 & 12.6 \\
\quad Recall    & \textbf{74.8} & 72.6 & 72.6 & 66.7 & 57.8 & 70.4 & 69.1 \\
\quad $F_{0.5}$ & \textbf{16.6} & 15.3 & 15.7 & 15.0 & 12.5 & 15.5 & 15.1 \\
\hline
Gemini-2.5-Pro & & & & & & & \\
\quad Precision & \textbf{17.9} & 17.5 & \textbf{17.9} & 17.3 & 15.5 & 17.5 & 17.2 \\
\quad Recall    & 60.7 & \textbf{62.2} & \textbf{62.2} & 57.0 & 51.1 & 60.7 & 59.0 \\
\quad $F_{0.5}$ & \textbf{20.9} & 20.4 & 20.8 & 20.1 & 18.0 & 20.4 & 20.1 \\
\hline
GPT-4o-mini & & & & & & & \\
\quad Precision & \textbf{35.2} & 26.5 & 27.5 & 26.5 & 31.9 & 29.4 & 29.5 \\
\quad Recall    & \textbf{14.1} & 6.7  & 8.2  & 9.6  & 11.1 & 11.1 & 10.1 \\
\quad $F_{0.5}$ & \textbf{27.1} & 16.6 & 18.6 & 19.6 & 23.2 & 22.1 & 21.2 \\
\hline
GPT-4o & & & & & & & \\
\quad Precision & 31.1 & 25.7 & \textbf{31.5} & 27.9 & 27.4 & 31.0 & 29.1 \\
\quad Recall    & 17.0 & 14.1 & 17.0 & 14.1 & 14.8 & \textbf{20.0} & 16.2 \\
\quad $F_{0.5}$ & 26.7 & 22.0 & 26.9 & 23.3 & 23.4 & \textbf{28.0} & 25.1 \\
\hline
GPT-5 & & & & & & & \\
\quad Precision & \textbf{63.6} & 33.3 & 50.0 & 58.3 & \textbf{63.6} & 53.3 & 53.7 \\
\quad Recall    & 5.2  & 2.2  & 2.2  & 5.2  & 5.2  & \textbf{5.9}  & 4.3 \\
\quad $F_{0.5}$ & 19.6 & 8.8  & 9.4  & 19.1 & 19.6 & \textbf{20.5} & 16.2 \\
\hline
o3 & & & & & & & \\
\quad Precision & 35.5 & 26.1 & 35.7 & 47.1 & \textbf{58.3} & 52.0 & 42.4 \\
\quad Recall    & 8.2  & 4.4  & 3.7  & 5.9  & \textbf{10.4} & 9.6  & 7.0 \\
\quad $F_{0.5}$ & 21.2 & 13.2 & 13.1 & 19.7 & \textbf{30.3} & 27.7 & 20.9 \\
\hline
\end{tabularx}

\label{tab:per_fold_results}
\end{table}
\FloatBarrier

\section{Ablation Analysis - Pre-anonymization Precision Rates}\label{app:pre-anonymization_rates}

As an additional ablation analysis, we evaluated the performance of two models on pre-anonymization inputs rather than the VCBench format. More precisely, we used the Processed JSON format, which hides the founder name field but applies no further anonymization measures. Table~\ref{tab:pre-anonymisation_rates} shows a substantial increase in predictive performance. Although both data contamination and the greater diversity of features could lead to this inflation, we suspect that the former contributes more given the scale of the disparity.

\begin{table}[h]
\centering
\caption{Average predictive performance of GPT-4o and o3 using the pre-anonymization format (Processed JSON), with performances on VCBench shown in brackets for comparison}
\setlength{\tabcolsep}{4pt}
\begin{tabular}{l c c c}
\hline
Model & Precision (\%)& Recall (\%)& $F_{0.5}$ (\%)\\
\hline
GPT-4o              & \textbf{45.8} (29.1) & \textbf{20.0} (16.2) & \textbf{36.4} (25.1) \\
o3                  & \textbf{51.7} (42.4) & \textbf{20.6} (7.0) & \textbf{39.6} (20.9) \\

\hline
\end{tabular}

\label{tab:pre-anonymisation_rates}
\end{table}

\end{document}